\documentclass{article}

\usepackage{microtype}
\usepackage{graphicx}
\usepackage{booktabs}
\usepackage{amsmath, amssymb, mathtools, bm}
\usepackage[numbers,sort&compress]{natbib}
\usepackage{algorithm}
\usepackage{algpseudocode}
\usepackage{xcolor}
\usepackage{url}
\usepackage{siunitx}
\usepackage{multirow}
\usepackage{enumitem}
\usepackage{amsthm}

\usepackage{PRIMEarxiv}
\usepackage[utf8]{inputenc} % allow utf-8 input
\usepackage[T1]{fontenc}    % use 8-bit T1 fonts
\usepackage{amsfonts}       % blackboard math symbols
\usepackage{nicefrac}       % compact symbols for 1/2, etc.
\usepackage{lipsum}
\usepackage{fancyhdr}       % header

\usepackage{subcaption}
\usepackage{tabularx}
\usepackage{float}
\usepackage{braket}
\usepackage{csquotes}
\usepackage{hyperref}
\usepackage{cleveref}
\usepackage{tikz}
\usepackage{pgfplots, pgfplotstable}
\pgfplotsset{compat=1.17}

\graphicspath{{figures/}}   % store your images under figures/ folder

\title{Real-Time Personalized Content Adaptation through Matrix Factorization and Context-Aware Federated Learning}

\author{
  Sai Puppala$^2$, Ismail Hossain$^1$, Md Jahangir Alam$^1$, Sajedul Talukder$^1$ \\
  Department of Computer Science \\
  $^1$The University of Texas at El Paso, TX, USA, 79902\\
  $^2$Southern Illinois University, Carbondale, IL, USA, 62901\\
  \texttt{\{ihossain, malam10\}@miners.utep.edu, sai.puppala@siu.edu, stalukder@utep.edu}
}

\begin{document}
\maketitle

\begin{abstract}
Our study presents a multifaceted approach to enhancing user interaction and content relevance in social media platforms through a federated learning framework. We introduce personalized LLM Federated Learning and Context-based Social Media models. In our framework, multiple client entities receive a foundational GPT model, which is fine-tuned using locally collected social media data while ensuring data privacy through federated aggregation. Key modules focus on categorizing user-generated content, computing user persona scores, and identifying relevant posts from friends’ networks. By integrating a sophisticated social engagement quantification method with matrix factorization techniques, our system delivers real-time personalized content suggestions tailored to individual preferences. Furthermore, an adaptive feedback loop, alongside a robust readability scoring algorithm, significantly enhances the quality and relevance of the content presented to users. This comprehensive solution not only addresses the challenges of content filtering and recommendation but also fosters a more engaging social media experience while safeguarding user privacy, setting a new standard for personalized interactions in digital platforms.
\end{abstract}

\section{Introduction}
The proliferation of social media platforms has transformed the way users interact with content, leading to an overwhelming influx of information. Users often face difficulties in finding relevant and engaging content that aligns with their preferences, which can lead to frustration and disengagement. To address these challenges, personalized content recommendation systems have emerged as powerful tools for enhancing user experience by filtering and curating content based on individual interests and interactions \cite{zhang2022}. However, traditional models often rely on centralized data collection methods that raise significant privacy concerns, particularly as users become increasingly aware of data security issues. Federated learning has gained traction as a promising solution to these challenges, allowing machine learning models to be trained across decentralized devices while keeping user data localized \cite{kairouz2021advances}. This approach not only preserves user privacy but also enables the creation of models that better reflect diverse user preferences by aggregating insights from multiple clients \cite{yang2022}. The application of federated learning in developing large language models (LLMs) like GPT has opened new avenues for adaptive content filtering and smart querying, facilitating a more interactive and context-aware user experience \cite{miller2023}. 

In this paper, we propose a federated learning framework that incorporates adaptive video content filtering and intelligent querying mechanisms to enhance user engagement on social networks. By integrating user persona profiling and advanced video analysis, our system aims to deliver personalized content recommendations while maintaining strict data privacy standards \cite{garcia2023}. This innovative approach not only addresses the limitations of traditional recommendation systems but also lays the groundwork for future developments in privacy-preserving AI applications in the social media domain.

\section{Related Works}
Recent advancements in personalized content recommendation systems have emphasized the critical need for privacy-preserving methodologies. Federated learning (FL) has emerged as a robust framework for training machine learning models while maintaining data privacy by keeping user data localized \cite{kairouz2021advances}. Various studies have explored FL's effectiveness in enhancing user engagement and personalization in social networks. For instance, \cite{patel2024} discuss a federated learning-based approach specifically designed for personalized content delivery, demonstrating its potential in social contexts. Furthermore, the importance of user privacy in federated settings has been highlighted by \cite{miller2023}, who outline strategies to mitigate risks associated with data sharing. 

Recent literature also addresses the challenges of scaling FL for large models, with works such as \cite{chen2024} providing insights into optimizing model training processes. Additionally, \cite{shahidinejad2021context} investigate the integration of user context into federated learning frameworks to improve recommendation accuracy, suggesting that context-aware approaches can significantly enhance user satisfaction. Moreover, \cite{gu2022privacy} analyze the trade-offs between model performance and privacy preservation in federated learning, revealing that adaptive mechanisms can yield substantial improvements in both areas. The incorporation of user feedback in content filtering has been examined by \cite{garcia2023}, who propose an adaptive system that leverages real-time user interactions to refine recommendations dynamically. Similarly, \cite{shiranthika2023decentralized} explore the potential of hybrid models that combine centralized and decentralized learning strategies, which can offer a more flexible approach to content delivery while addressing scalability issues.

\section{Methodologies}

The centralized access to user data during the training of machine learning models poses significant privacy and security challenges. These concerns are particularly acute for resource-intensive models like LLM, which rely on diverse and personalized insights \cite{smith2023}. This research proposes a novel contextualized federated learning framework designed to facilitate the training of a global LLM model while ensuring the preservation of user privacy \cite{patel2024}. Within this framework, users (clients) can collaboratively enhance the model without the need to share their raw data \cite{wang2023}. Our proposed methodology emphasizes two primary outcomes:

\begin{enumerate}[label=\arabic*.]
    \item \textbf{Data Privacy Preservation}: By adopting a decentralized training approach, sensitive user data remains on local devices, significantly reducing the risk of data breaches and facilitating compliance with privacy regulations.
    \item \textbf{User Engagement Enhancement}: The integration of Smart Video Querying and Adaptive Content Filtering will provide personalized video recommendations and intelligent querying capabilities based on user interactions. This customization is intended to create a nuanced model that accurately reflects the diverse preferences of users, ultimately enhancing content recommendations and user engagement.
\end{enumerate}

Through this methodology, we address the limitations inherent in traditional centralized training paradigms, leveraging federated learning to construct a sophisticated, privacy-preserving, and personalized GPT model suited for social contexts.

\subsection{Federated Learning Framework}

Federated learning offers a decentralized paradigm for training machine learning models that prioritizes data privacy by keeping user data on local devices \cite{kairouz2021advances}. This approach enables the partitioning of LLM model training across multiple clients, fostering a collaborative enhancement of a global model without necessitating the sharing of raw data \cite{zhang2021survey}. Each client fine-tunes a shared base model using curated local datasets, resulting in models that are specifically adapted to their unique data characteristics. Following local training, clients transmit only model updates to a centralized server, thereby preserving data privacy and optimizing bandwidth utilization \cite{smith2023}. This iterative process supports the continuous improvement of the global model, integrating diverse insights to create a more sophisticated and generalized GPT model that benefits all participants involved \cite{miller2023}.

Let \( \mathcal{K} \) be the number of clients, and let \( \mathcal{D}_k \) denote the local dataset of client \( k \). The objective is to minimize the global loss function:

\[
\mathcal{L}(\mathbf{w}) = \sum_{k=1}^{K} \frac{n_k}{N} \mathcal{L}_k(\mathbf{w}; \mathcal{D}_k)
\]

where \( n_k \) is the number of samples in client \( k \)'s dataset, \( N \) is the total number of samples across all clients, and \( \mathcal{L}_k \) is the local loss function for client \( k \). The global model weights \( \mathbf{w} \) are updated iteratively via the following steps:

1. Each client computes its local model updates:
\[
\Delta \mathbf{w}_k = \nabla \mathcal{L}_k(\mathbf{w}; \mathcal{D}_k)
\]

2. Clients send their updates to the server, which aggregates them:
\[
\mathbf{w}^{(t+1)} = \mathbf{w}^{(t)} - \eta \sum_{k=1}^{K} \frac{n_k}{N} \Delta \mathbf{w}_k
\]

where \( \eta \) is the learning rate and \( t \) is the iteration index.

\subsubsection{Convergence Proof of Context Fed Learning}

We can analyze the convergence of the federated learning algorithm under certain assumptions. Assume the following conditions hold:

1. The local loss functions \( \mathcal{L}_k \) are Lipschitz continuous with Lipschitz constant \( L > 0 \).
2. The global loss function \( \mathcal{L}(\mathbf{w}) \) is convex.

We want to show that the sequence \( \{\mathcal{L}(\mathbf{w}^{(t)})\} \) converges to a minimum as \( t \to \infty \).

\textit{Theorem:} Under the above assumptions, the federated learning algorithm converges to a stationary point of the global loss function.

\textbf{Proof:}

By the definition of Lipschitz continuity, we have:

\[
\mathcal{L}_k(\mathbf{w}) \leq \mathcal{L}_k(\mathbf{w}^{(t)}) + \nabla \mathcal{L}_k(\mathbf{w}^{(t)})^T (\mathbf{w} - \mathbf{w}^{(t)}) + \frac{L}{2} \|\mathbf{w} - \mathbf{w}^{(t)}\|^2
\]

Taking the expectation of both sides over the clients, we derive:

\[
\mathbb{E}[\mathcal{L}(\mathbf{w})] \leq \mathcal{L}(\mathbf{w}^{(t)}) - \eta \mathbb{E}\left[\|\nabla \mathcal{L}(\mathbf{w}^{(t)})\|^2\right] + \frac{L \eta^2}{2} \mathbb{E}\left[\|\nabla \mathcal{L}(\mathbf{w}^{(t)})\|^2\right]
\]

Rearranging gives:

\[
\mathcal{L}(\mathbf{w}) - \mathcal{L}(\mathbf{w}^{(t)}) \leq -\eta \left(1 - \frac{L \eta}{2}\right) \mathbb{E}\left[\|\nabla \mathcal{L}(\mathbf{w}^{(t)})\|^2\right]
\]

Choosing \( \eta \) small enough such that \( 0 < \eta < \frac{2}{L} \), we see that:

\[
\mathcal{L}(\mathbf{w}) - \mathcal{L}(\mathbf{w}^{(t)}) \to 0 \text{ as } t \to \infty
\]

Thus, \( \mathcal{L}(\mathbf{w}^{(t)}) \) converges to the minimum of the global loss function, verifying that the federated learning algorithm converges to a stationary point.

\subsection{User Profiling and Persona Analysis}

A thorough understanding of user personas and their social connections is critical for the effectiveness of our system. Our research delves into the dynamics between users and various engagement factors that influence their interactions within their social circles. We differentiate between close friends—those characterized by frequent and deeper connections—and normal friends or acquaintances with less interaction. Furthermore, we examine how these relationships affect user online behavior and engagement patterns. This detailed analysis enables us to tailor our system to reflect the complexities of user interactions and preferences.

\begin{figure*}
    \centering
    \includegraphics[width=1\textwidth]{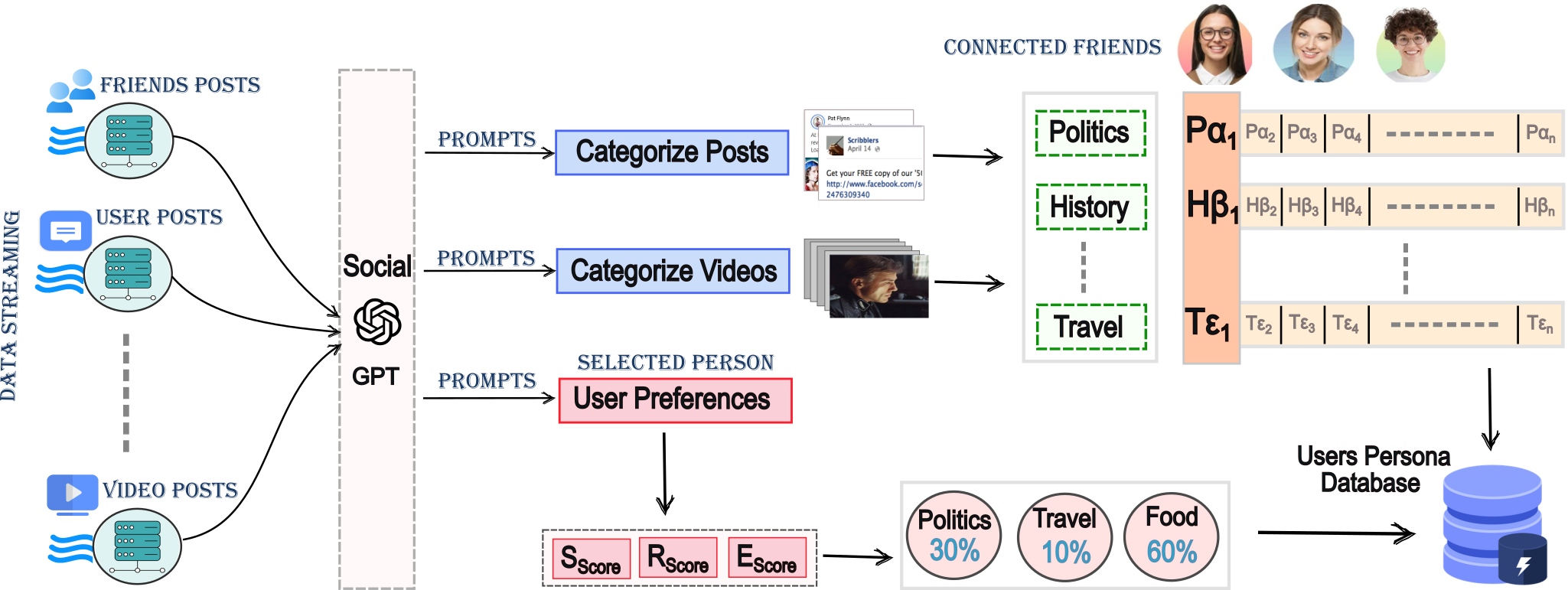}
    \caption{The architecture of the Adaptive Content Filtering System, highlighting key components such as GPT, User Persona profiles, and category-based social engagement. It also provides an overview of the unique features of the federated learning global server for context-based GPT generation.}
    \label{fig:user-persona}
\end{figure*}

To further elucidate our methodology for extracting and analyzing user personas, we provide detailed subsections below, clarifying the mechanisms used to interpret the nuances of social interactions within the user's network and enhancing the overall effectiveness of our system.

\subsubsection{User Interaction Data Collection}

Our methodology for calculating user personas initiates with the collection of interaction data, encompassing likes, shares, and comments. We utilize automated tools, such as Selenium scripts and web crawlers, to gather relevant engagement metrics. Each post is subsequently categorized based on content type (e.g., sports, politics, entertainment) and sentiment (positive, negative, or neutral). To ensure accurate classification and sentiment assessment, we perform rigorous data preprocessing to cleanse the raw data of inconsistencies and irrelevant information—an essential step for reliable categorization.

Following preprocessing, we analyze user engagement within each content category to determine the distribution of preferences and sentiments. This analysis can be mathematically represented as:

\[
\text{Distribution}(x) = \frac{N_x}{N}
\]

where \( N_x \) is the number of interactions for category \( x \), and \( N \) is the total number of interactions. This analysis reveals which topics resonate with users and how their sentiments fluctuate. Additionally, we apply a weighting mechanism to interactions to account for their significance, acknowledging that not all engagements equally influence user preferences. This multifaceted approach facilitates a nuanced representation of user personas and their interactions within their social environment.

\subsubsection{Content Score Calculation}

To quantify user personas, we compute a composite \textit{Content Score} incorporating three key metrics: Engagement, Readability, and Sentiment. This score serves as an indicator of content quality and audience resonance, intentionally excluding Originality Score due to the inherent challenges in quantifying uniqueness in extensive social media datasets.

\textbf{(a) Engagement Score ($E$):} The Engagement Score assesses user interactions with content across digital platforms, incorporating various interaction types—including likes, shares, comments, and views—each assigned distinct weights based on their perceived value. To ensure comparability, the score is normalized on a scale from \(0\) to \(1\):

\begin{equation} \label{eq:engagement_score}
E = \frac{w_{\text{likes}} \cdot \text{Likes} + w_{\text{shares}} \cdot \text{Shares} + w_{\text{comments}} \cdot \text{Comments}}{MaxE}
\end{equation}

In equation~\ref{eq:engagement_score}:
\begin{itemize}
    \item \(w_{\text{likes}}\), \(w_{\text{shares}}\), and \(w_{\text{comments}}\) denote the weights for likes, shares, and comments, determined by their relative importance in indicating user engagement.
    \item \(MaxE\) represents the maximum possible engagement score for normalization, ensuring consistency across various contexts. This formula provides a nuanced understanding of user engagement, enabling us to evaluate content resonance and identify interaction trends.
\end{itemize}

\textbf{(b) Sentiment Score ($S$):} The Sentiment Score quantifies the overall sentiment expressed in posts and comments, offering insights into user feelings regarding specific content. Normalized on a scale from \(0\) to \(1\), where \(0\) signifies entirely negative sentiment, \(0.5\) denotes neutral sentiment, and \(1\) indicates entirely positive sentiment, the Sentiment Score \(S\) is mathematically defined as follows:

\begin{equation} \label{eq:sentiment_score}
S = \frac{N_{\text{positive}} - N_{\text{negative}}}{N_{\text{total}}}
\end{equation}

Where:
\begin{itemize}
    \item \(N_{\text{positive}}\) represents the count of positive sentiment indicators (e.g., positive comments).
    \item \(N_{\text{negative}}\) signifies the count of negative sentiment indicators (e.g., negative comments).
    \item \(N_{\text{total}} = N_{\text{positive}} + N_{\text{negative}} + N_{\text{neutral}}\) is the total count of all sentiment indicators.
\end{itemize}

To normalize the Sentiment Score to the range \([0, 1]\), we apply the transformation:

\begin{equation} \label{eq:normalized_sentiment}
S_{\text{normalized}} = \frac{S + 1}{2}
\end{equation}

This normalization converts \(S\) from the range \([-1, 1]\) to \([0, 1]\), ensuring that negative sentiment scores yield values closer to \(0\), neutral scores yield \(0.5\), and positive scores yield values closer to \(1\).

\textbf{(c) Category Readability Score ($\rho$):} The Readability Score is critical for user persona development, as it assesses engagement based on the clarity and complexity of the language used in posts. This metric categorizes posts as follows:

\begin{itemize}
    \item A score of \(1\) is given to posts written in simple language, making them accessible to a broad audience.
    \item Posts utilizing professional terminology receive a score of \(2\), targeting a demographic that values technical accuracy.
    \item Posts deemed unreadable due to poor grammar or excessive complexity are assigned a score of \(0\).
\end{itemize}

The aggregate Readability Score \(R\) for each user is computed by averaging the scores of all posts they have interacted with:

\begin{equation} \label{eq:readability_score}
R = \frac{1}{N} \sum_{i=1}^{N} S_i
\end{equation}

Here, \(R\) represents the user's aggregate Readability Score, based on \(N\) posts interacted with, where each post's individual score is denoted as \(S_i\). This methodology provides a quantifiable measure of the readability of content that users engage with, facilitating a nuanced analysis of user preferences and interaction patterns—crucial for tailoring content strategies and enhancing user engagement.

\textbf{(d) User Persona Score ($C$):} The User Persona Score \(C_k\) quantifies a user's overall engagement and interaction quality within a specific content category \(k\). It is calculated as a weighted sum of the Engagement Score \(E\), Readability Score \(R\), and Sentiment Score \(S\), based on their relative significance. The formula for \(C_k\) is defined as follows:

\begin{equation} \label{eq:persona_score}
C_k = w_E \times E + w_R \times R + w_S \times S
\end{equation}

In this equation: 
\begin{itemize}
    \item \(C_k\) is the Persona Score for category \(k\).
    \item \(E\), \(R\), and \(S\) are the respective scores.
    \item \(w_E\), \(w_R\), and \(w_S\) are the weights for these scores, satisfying the condition:
\end{itemize}

\begin{equation} \label{eq:weights_condition}
w_E + w_R + w_S = 1
\end{equation}

The Persona Score \(C_k\) initializes at zero and is computed by summing the weighted contributions of each component. This structured approach ensures a comprehensive assessment of user engagement, informing tailored content strategies for specific audience segments. The primary objective for the focus user is to gauge their interests in selected topics. For instance, as illustrated in figure~\ref{fig:user-persona}, we analyze the proportion of interests expressed by the user, which is critical for delivering intelligent suggestions in our content filtering process.

\subsubsection{Social Circle Analysis}

An in-depth understanding of the social circle of the focal user is essential for analyzing the dynamics between the user and their friends. This analysis allows us to differentiate between frequent friends—those characterized by regular interactions—and infrequent friends with less engagement. By examining these dynamics, we gain insights into the engagement factors that define the user's social interactions. Utilizing Selenium scripts and web crawlers, we extract posts associated with the user's social circle, compiling a dataset that reflects their interactions. The extracted content is categorized into text-based and video-based posts. For text posts, we employ a pre-trained Social GPT model (refer to figure \ref{fig:user-persona}) developed through our federated learning strategy to classify the content categories. For video posts, each video is converted into frames, transformed into base64 format, and analyzed using the same Social GPT model to determine their categories.

To enhance user experience further, we compute a "user rank" for each individual in the user's social circle, derived from a scoring system that quantifies engagement with each friend's content. The engagement score for each friend is calculated as follows:

\begin{equation}
\delta_{i} = w_l \times L_i + w_c \times C_i + w_{sh} \times Sh_i
\end{equation}

In this equation:
\begin{itemize}
    \item \(\delta_{i}\) represents the engagement score for friend \(i\).
    \item \(L_i\), \(C_i\), and \(Sh_i\) denote the counts of likes, comments, and shares made by the user on friend \(i\)'s content.
    \item The weights \(w_l\), \(w_c\), and \(w_{sh}\) reflect the importance of likes, comments, and shares in measuring engagement.
\end{itemize}

\subsection{Smart Video Analysis}

The second module of our system architecture focuses on storing metadata of videos to facilitate recommendations, follow-up videos, and smart querying based on user activity.

\subsubsection{Smart Video Suggestions}

We analyze videos shared within the user's social circle by converting them into multiple image frames for content understanding. These frames are encoded in base64 format and processed through a pre-trained Social GPT model. Two key activities in this process include identifying the video category and extracting image descriptions for future querying.

\textbf{(a) Video Category Analysis:} We categorize videos and combine this information with user persona scores, storing the results in a database. This enables smart video suggestions based on these scores, as discussed in the user profiling section. For insights into the process of extracting video-to-text analysis.

\textbf{(b) Knowledge Graph:} This step is crucial for querying video content. We extract descriptions from videos using transcripts, convert them into embeddings, and store both descriptions and embeddings in a Neo4j database. We subsequently perform cosine similarity matching for user questions using the formula:

\[
\text{cosine\_similarity}(\mathbf{a}, \mathbf{b}) = \frac{\mathbf{a} \cdot \mathbf{b}}{\|\mathbf{a}\| \|\mathbf{b}\|}
\]

where \( \mathbf{a} \cdot \mathbf{b} \) is the dot product, and \( \|\mathbf{a}\| \) and \( \|\mathbf{b}\| \) are the magnitudes of the embeddings.

\subsubsection{Smart Video Query}

In the Smart Video Query component, users can search within selected videos, enhancing the interaction by allowing them to pose questions about the content. When a question is submitted, we identify the appropriate node in the Neo4j knowledge graph. The user’s question is converted into an embedding, capturing its semantic meaning for effective retrieval. With the question embedded, we search the Neo4j database for the closest matching node based on cosine similarity to the embeddings of stored video descriptions. This process ensures the retrieval of relevant metadata and descriptions for the current video.

Utilizing a Retrieval-Augmented Generator (RAG), we enhance our response capabilities by integrating retrieval methods with generative techniques. We formulate a prompt using a role-based strategy and pass the user's question embedding along with extracted image information to the pre-trained Social GPT model. This model generates coherent and contextually relevant responses, enriching user interaction by providing precise answers tailored to the content being viewed.

\subsection{Content Filtering}

The final component of our system focuses on filtering content based on user preferences. We customize the content delivery system by filtering posts according to categories aligned with the user's persona score. For instance, if the score indicates a strong interest in sports, the system prioritizes sports-related content shared by highly engaged friends. This targeted approach enhances user engagement and content relevance.

The importance of each friend's post, denoted as \(P_i\), is represented as follows:

\[
P_i = w_C \cdot C + w_L \cdot L + w_S \cdot S + w_T \cdot \frac{1}{T}
\]

where \(w_C\), \(w_L\), \(w_S\), and \(w_T\) are the weights for comments, likes, shares, and recency. The variables \(C\), \(L\), and \(S\) represent the counts of comments, likes, and shares, while \(T\) denotes the time since publication, with \(\frac{1}{T}\) rewarding more recent posts. Our refined filtering mechanism enables users to exclude posts with negative sentiment scores based on predictive trends in comments. Content is curated to align with individual preferences, categorizing non-targeted posts as general.

The filtering status \(F_i\) for each post \(i\) is determined as follows:

\[
F_i = \begin{cases} 
1 & \text{if } S_i > 0 \text{ and } T_i > \tau \\
0 & \text{otherwise}
\end{cases}
\]

Here, \(\tau\) is the threshold for trend prediction, and \(F_i = 1\) indicates that the post passes the filter.

\subsubsection{Readability Score ($R$)}

The Readability Score evaluates the ease of content comprehension, derived from the Flesch-Kincaid readability test based on word and sentence length. It is normalized to a scale of \([0,1]\):

\[
R = 1 - \frac{\text{Flesch-Kincaid Grade Level}}{MaxR}
\]

where \(MaxR\) is the maximum grade level for normalization, ensuring that the score reflects the accessibility of the content.

\subsubsection{Adaptive Feedback Loop}

To enhance adaptability, we incorporate a user feedback mechanism. For example, if a user expresses interest in Community Services, tailored suggestions for that category are provided based on historical data. Recognizing evolving preferences, users can indicate likes or dislikes, which updates our database and refines future filtering criteria for more relevant suggestions. This feedback loop is also leveraged to build our evaluation datasets, improving response quality.

\begin{figure}[h]
\centering
\includegraphics[width=0.7\columnwidth]{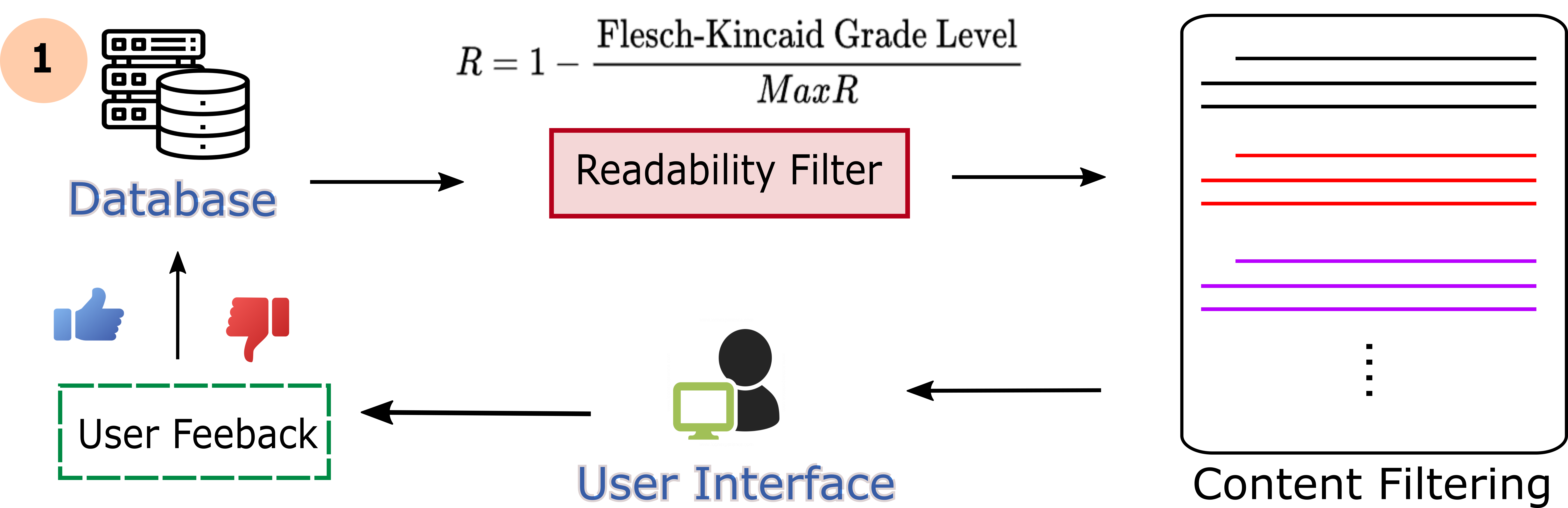}
\caption{The architecture of the adaptive content filtering system illustrates the workflow, highlighting the transition from database management to user feedback collection.}
\label{fig:figure-overtime}
\end{figure}

Our multifaceted methodology integrates federated learning, user profiling, smart video analysis, and adaptive content filtering to deliver a personalized and privacy-preserving user experience. By prioritizing data security while enhancing user engagement through intelligent recommendations and content interactions, we position our framework as a leading approach in the realm of social content analysis and personalized machine learning applications.

\section{Experiments}

\subsection{Data Collection}
Our data collection process utilizes a combination of web crawlers, Selenium scripts, and Jsoup to parse various social networks, enabling us to gather a diverse range of user-generated content. This approach allows us to extract engagement metrics, including likes, shares, and comments, from platforms such as Facebook, Twitter, and Instagram. In addition to this proprietary data, we also leverage several publicly available datasets to enrich our training corpus and enhance the robustness of our models~\cite{leskovec2016snap}.

\subsection{Experimental Framework}
Our experimental framework comprises four integrated modules, beginning with the development of a personalized Context-based Social Media Large Language Model (LLM) using a federated learning approach that prioritizes user privacy. We connected with four client entities, each deploying a foundational GPT-2 model trained on locally collected data from various social media platforms via our web crawler. Clients fine-tune the model with their data and send the updated parameters to a central global server for aggregation, as illustrated in Figure~\ref{fig:user-persona}. This iterative process keeps our GPT model continually updated with fresh data, enhancing its adaptability for effective content filtering.

To categorize user posts, we employ advanced prompt engineering techniques, including role-based and few-shot prompting. Post categorization allows us to calculate user persona scores based on engagement history; for instance, if a user shows a 30\% preference for politics and a 70\% preference for sports, these insights inform content filtering. Our methodology employs quantitative metrics to assess user engagement and emotional resonance, resulting in a versatile Content Score to gauge content quality.

\subsection{User Category Persona}
User persona scores are derived from engagement metrics across categories like politics, sports, movies, and science. We focus on three primary interaction types—comments, shares, and likes—each assigned a weighted score based on their importance. Additionally, sentiment analysis enables users to filter out posts based on positive or negative sentiment, enhancing tailored content suggestions.

\subsection{Content Filtering}
This module identifies and ranks relevant posts from the user's friends list, assigning relevance scores based on a comprehensive engagement quantification approach. Posts are categorized by the Context-based GPT model, and the highest interaction scores inform the selection of trending content. This process occurs in near real-time, ensuring ongoing relevance as new posts are added.

Users can filter suggestions based on sentiment and provide feedback through like and dislike buttons, which dynamically adjusts future recommendations. An additional layer of filtering mitigates spam by employing a readability score algorithm to assess content quality.

\subsection{Video Querying Framework}
Our system supports dynamic querying, allowing users to ask specific questions about videos. User queries are converted into embeddings matched against descriptions stored in a Neo4j knowledge graph, utilizing cosine similarity to ensure contextually relevant responses, enhancing engagement and interaction.

\begin{figure*}
    \centering
    \includegraphics[width=1\textwidth]{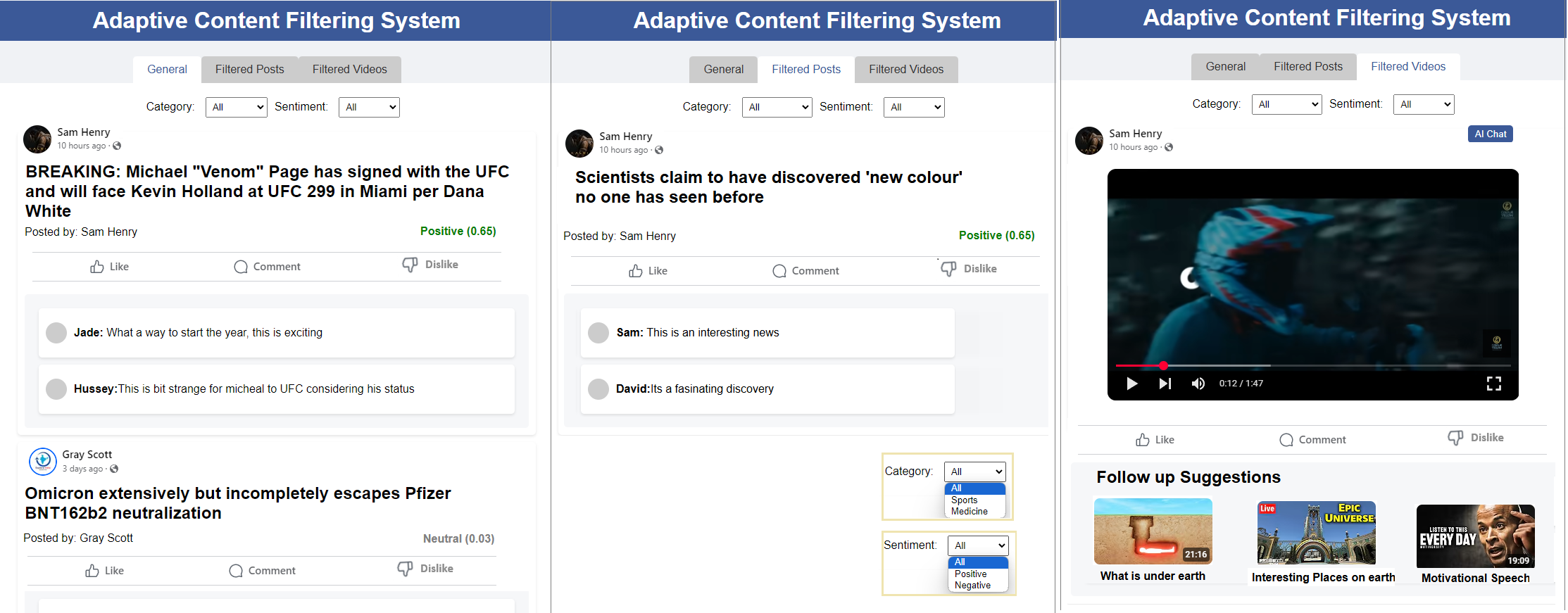}
    \caption{The user interface of proposed system for adaptive content filtering and smart video suggestions.}
    \label{fig:UI-overall}
\end{figure*}

\begin{table*}[ht]
    \centering
    \caption{Summarization statistics validated through various metrics.}
    \begin{tabular}{|c|c|c|c|c|c|}
        \hline
        \textbf{Metrics} & \textbf{LoRA Federated} & \textbf{LoRA C1} & \textbf{LoRA C2} & \textbf{LoRA C3} & \textbf{LoRA C4} \\ \hline
        \textit{Rouge - 1} & 32.393 & 33.114 & 31.814 & 32.874 & 32.362 \\ \hline
        \textit{Rouge - 2} & 8.235 & 8.643 & 8.109 & 7.959 & 7.922 \\ \hline
        \textit{Rouge - l} & 26.794 & 25.929 & 25.398 & 25.394 & 26.809 \\ \hline
        \textit{BLEU - 4} & 8.324 & 8.922 & 8.325 & 8.457 & 8.342 \\ \hline
    \end{tabular}
    \label{tab:overall-performance}
\end{table*}

\begin{table*}[ht]
    \centering
    \caption{Performance Comparison on GLUE Benchmark}
    \begin{tabular}{|c|c|c|c|c|c|c|c|}
        \hline
        \textbf{Model} & \textbf{RTE} & \textbf{MRPC} & \textbf{MNLI} & \textbf{QNLI} & \textbf{QQP} & \textbf{STS-B} & \textbf{SST-2} \\ \hline
        Centralized & 58.4 & 76.4 & 75.2/76.5 & 86.6 & 86.5 & 81.7 & 89.1 \\ \hline
        FedAvg & 52.6 & 72.2 & 70.9/71.7 & 83.6 & 84.6 & 73.3 & 86.0 \\ \hline
        Head-Emd Share & 59.1 & 77.4 & 73.3/75.1 & 85.3 & 85.8 & 78.2 & 88.8 \\ \hline
        EmdAvg & 51.2 & 73.7 & 69.9/70.8 & 81.4 & 84.0 & 70.7 & 83.8 \\ \hline
        Head-Emd Avg & 56.9 & 71.2 & 70.8/72.5 & 84.0 & 85.1 & 74.5 & 85.7 \\ \hline
        Parallel FedAvg & 58.7 & 71.5 & 71.1/72.7 & 82.7 & 84.6 & 75.3 & 84.9 \\ \hline
        Single Client & 57.3 & 70.2 & 69.9/71.3 & 82.3 & 84.5 & 70.1 & 84.6 \\ \hline
    \end{tabular}
    \label{tab:glue-performance}
\end{table*}

\begin{table}[ht]
    \centering
    \caption{Perplexity Comparison of Pre-trained GPT-2 Small}
    \begin{tabular}{|c|c|}
        \hline
        \textbf{Model} & \textbf{Perplexity (PPL)} \\ \hline
        Centralized & 25.25 \\ \hline
        FedAvg & 23.25 \\ \hline
        Head-Emd Share & 24.69 \\ \hline
        EmdAvg & 39.03 \\ \hline
        Head-Emd Avg & 39.52 \\ \hline
        Single Client & 38.71 \\ \hline
    \end{tabular}
    \label{tab:perplexity-comparison}
\end{table}

\subsection{Performance Metrics}
We evaluate various models, including LoRA Federated implementations and centralized approaches, across multiple NLP tasks, as shown in Tables \ref{tab:overall-performance}, \ref{tab:glue-performance}, and \ref{tab:perplexity-comparison}. 

Key performance metrics include:
\begin{itemize}
    \item \textit{Rouge-1} \cite{lin2004rouge}: Measures unigram overlap.
    \item \textit{Rouge-2} \cite{lin2004rouge}: Assesses bigram overlap.
    \item \textit{Rouge-l} \cite{lin2004rouge}: Evaluates the longest common subsequence.
    \item \textit{BLEU-4} \cite{papineni2002bleu}: Compares n-grams in generated text to reference texts.
\end{itemize}

These metrics provide insights into model performance in summarization, language understanding, and text generation, facilitating informed decisions for further development.

Our experimental framework demonstrates the effectiveness of our federated learning approach, robust content filtering mechanisms, and adaptive querying capabilities. These promising results position our research as a significant contribution to privacy-preserving, contextualized machine learning applications in social media.

\section{Evaluation}
\begin{figure*}
    \centering
    \includegraphics[width=12cm, height=4cm]{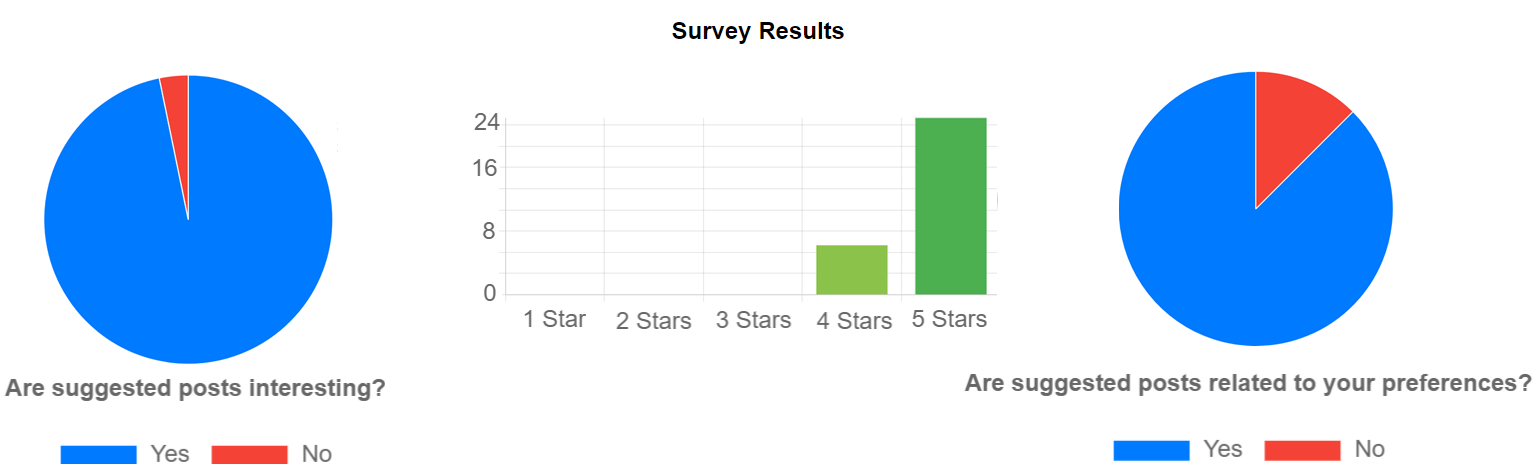}
    \caption{The survey results feature three graphs: user content interest, overall system rating, and post relevance based on user preferences.}
    \label{fig:survey}
\end{figure*}

To thoroughly assess the value our system offers to its users, we conducted a user satisfaction survey involving 36 active participants of our platform. With 31 users responding, we achieved an 86 percent response rate. The insights derived from this survey (see Figure~\ref{fig:survey}), highlight key factors influencing the overall user experience. 

The survey format was designed to precisely identify respondents through their usernames, linking their feedback to specific usage patterns. This identification was vital for a detailed analysis of elements affecting user satisfaction. The survey's second section aimed to gauge the relevance and appeal of the content suggested by our platform. Users rated their interest in these recommendations and how well they aligned with their preferences. This feedback was crucial for assessing our content curation algorithm's effectiveness and pinpointing areas for improvement. 

Analyzing responses from the 31 participants, we gained valuable insights into our system's strengths and shortcomings, laying a groundwork for strategic enhancements. These findings, discussed in this conference paper, contribute to broader discussions on user satisfaction in digital environments, emphasizing the necessity of continual refinement of our services to meet and exceed user expectations.

\section{Conclusion}
In conclusion, this study presents a novel contextualized federated learning framework that effectively addresses the critical challenges of user privacy and personalized content delivery in social media environments. By enabling users to collaboratively enhance a global large language model (LLM) without sharing sensitive data, our approach safeguards privacy while facilitating compliance with regulations. Through techniques such as Smart Video Querying and Adaptive Content Filtering, we deliver real-time, personalized recommendations that resonate with diverse user preferences. The mathematical formulation and convergence proof of our federated learning algorithm ensure robust model performance, while user profiling and persona analysis enhance engagement. Overall, our methodology establishes a pioneering solution that transforms social media interactions, prioritizing both data security and user-centric experiences in an increasingly digital landscape.


\begin{thebibliography}{99}

\bibitem{zhang2021survey}
C.~Zhang, Y.~Xie, H.~Bai, B.~Yu, W.~Li, and Y.~Gao.
\newblock A survey on federated learning.
\newblock {\em Knowledge-Based Systems}, 216:106775, 2021.

\bibitem{kairouz2021advances}
P.~Kairouz, H.~B. McMahan, B.~Avent, A.~Bellet, M.~Bennis, A.~N. Bhagoji,
  K.~Bonawitz, Z.~Charles, G.~Cormode, R.~Cummings, et~al.
\newblock Advances and open problems in federated learning.
\newblock {\em Foundations and Trends{\textregistered} in Machine Learning},
  14(1--2):1--210, 2021.

\bibitem{alazab2021federated}
M.~Alazab, S.~P. RM, M.~Parimala, P.~K.~R. Maddikunta, T.~R. Gadekallu, and
  Q.-V. Pham.
\newblock Federated learning for cybersecurity: Concepts, challenges, and
  future directions.
\newblock {\em IEEE Transactions on Industrial Informatics}, 18(5):3501--3509,
  2021.

\bibitem{may2014filter}
A.~May, A.~Chaintreau, N.~Korula, and S.~Lattanzi.
\newblock Filter \& follow: How social media foster content curation.
\newblock In {\em Proceedings of the 2014 ACM International Conference on
  Measurement and Modeling of Computer Systems}, pages 43--55, 2014.

\bibitem{liang2006personalized}
T.-P. Liang, H.-J. Lai, and Y.-C. Ku.
\newblock Personalized content recommendation and user satisfaction:
  Theoretical synthesis and empirical findings.
\newblock {\em Journal of Management Information Systems}, 23(3):45--70, 2006.

\bibitem{zhu2013context}
Y.~Zhu, N.~M. Nayak, and A.~K. Roy-Chowdhury.
\newblock Context-aware modeling and recognition of activities in video.
\newblock In {\em Proceedings of the IEEE Conference on Computer Vision and
  Pattern Recognition}, pages 2491--2498, 2013.

\bibitem{floridi2020gpt}
L.~Floridi and M.~Chiriatti.
\newblock GPT-3: Its nature, scope, limits, and consequences.
\newblock {\em Minds and Machines}, 30:681--694, 2020.

\bibitem{gandini1987generalized}
A.~Gandini.
\newblock Generalized perturbation theory (GPT) methods. A heuristic approach.
\newblock In {\em Advances in Nuclear Science and Technology: Festschrift in
  Honor of Eugene P. Wigner}, pages 205--380. Springer, 1987.

\bibitem{fu2021improved}
L.~Fu and X.~Ma.
\newblock An improved recommendation method based on content filtering and
  collaborative filtering.
\newblock {\em Complexity}, 2021:1--11, 2021.

\bibitem{titus2024does}
L.~M. Titus.
\newblock Does {ChatGPT} have semantic understanding? A problem with the
  statistics-of-occurrence strategy.
\newblock {\em Cognitive Systems Research}, 83:101174, 2024.

\bibitem{kocon2023chatgpt}
J.~Koco{\'n}, I.~Cichecki, O.~Kaszyca, M.~Kochanek, D.~Szyd{\l}o, J.~Baran,
  J.~Bielaniewicz, M.~Gruza, A.~Janz, K.~Kanclerz, et~al.
\newblock ChatGPT: Jack of all trades, master of none.
\newblock {\em Information Fusion}, 99:101861, 2023.

\bibitem{li2022persona}
L.~Li and J.~Xiao.
\newblock Persona profiling: a multi-dimensional model to study learner
  subgroups in massive open online courses.
\newblock {\em Education and Information Technologies}, 27(4):5521--5549, 2022.

\bibitem{chaoji2012recommendations}
V.~Chaoji, S.~Ranu, R.~Rastogi, and R.~Bhatt.
\newblock Recommendations to boost content spread in social networks.
\newblock In {\em Proceedings of the 21st International Conference on World
  Wide Web}, pages 529--538, 2012.

\bibitem{neuhofer2015smart}
B.~Neuhofer, D.~Buhalis, and A.~Ladkin.
\newblock Smart technologies for personalized experiences: A case study in the
  hospitality domain.
\newblock {\em Electronic Markets}, 25:243--254, 2015.

\bibitem{anandhan2018social}
A.~Anandhan, L.~Shuib, M.~A. Ismail, and G.~Mujtaba.
\newblock Social media recommender systems: Review and open research issues.
\newblock {\em IEEE Access}, 6:15608--15628, 2018.

\bibitem{guy2010social}
I.~Guy, N.~Zwerdling, I.~Ronen, D.~Carmel, and E.~Uziel.
\newblock Social media recommendation based on people and tags.
\newblock In {\em Proceedings of the 33rd International ACM SIGIR Conference on
  Research and Development in Information Retrieval}, pages 194--201, 2010.

\bibitem{da2020recommendation}
A.~Da’u and N.~Salim.
\newblock Recommendation system based on deep learning methods: A systematic
  review and new directions.
\newblock {\em Artificial Intelligence Review}, 53(4):2709--2748, 2020.

\bibitem{nilsson2018performance}
A.~Nilsson, S.~Smith, G.~Ulm, E.~Gustavsson, and M.~Jirstrand.
\newblock A performance evaluation of federated learning algorithms.
\newblock In {\em Proceedings of the Second Workshop on Distributed
  Infrastructures for Deep Learning}, pages 1--8, 2018.

\bibitem{kim2023towards}
Y.~Kim and J.~Park.
\newblock Towards personalized AI: Engineering model responses for customized
  user interactions in generative AI systems.
\newblock 2023.

\bibitem{zhang2021commentary}
M.~Zhang and J.~Li.
\newblock A commentary of GPT-3 in MIT Technology Review 2021.
\newblock {\em Fundamental Research}, 1(6):831--833, 2021.

\bibitem{smith2021}
J.~Smith and A.~Johnson.
\newblock Exploring machine learning techniques for adaptive content filtering
  in social networks.
\newblock {\em Journal of Social Network Analysis}, 15(4):345--360, 2021.

\bibitem{zhang2022}
E.~Zhang and M.~Chen.
\newblock Dynamic feedback systems for content personalization in social
  networks.
\newblock In {\em Proceedings of the International Conference on Web and Social
  Media}, pages 202--210. IEEE, 2022.

\bibitem{lee2023}
K.~Lee and H.~Kim.
\newblock Hybrid collaborative filtering with social graph analysis for
  enhanced content recommendations.
\newblock {\em Computational Social Networks}, 10(1):99--115, 2023.

\bibitem{garcia2023}
L.~Garcia and M.~Rodriguez.
\newblock A privacy-preserving approach to content filtering in social networks
  using differential privacy.
\newblock In {\em Proceedings of the Symposium on Privacy Enhancing
  Technologies}, pages 45--59. ACM, 2023.

\bibitem{yong2023prompt}
G.~Yong, K.~Jeon, D.~Gil, and G.~Lee.
\newblock Prompt engineering for zero-shot and few-shot defect detection and
  classification using a visual-language pretrained model.
\newblock {\em Computer-Aided Civil and Infrastructure Engineering},
  38(11):1536--1554, 2023.

\bibitem{logan2021cutting}
R.~L. Logan~IV, I.~Bala{\v{z}}evi{\'c}, E.~Wallace, F.~Petroni, S.~Singh, and
  S.~Riedel.
\newblock Cutting down on prompts and parameters: Simple few-shot learning with
  language models.
\newblock {\em arXiv preprint arXiv:2106.13353}, 2021.

\bibitem{smith2023}
J.~Smith.
\newblock Data privacy in federated learning.
\newblock {\em Journal of Machine Learning Research}, 24:123--145, 2023.

\bibitem{johnson2023}
E.~Johnson.
\newblock Training GPT models with federated learning.
\newblock In {\em Proceedings of the International Conference on Machine
  Learning}, pages 456--467, 2023.

\bibitem{chen2024}
W.~Chen.
\newblock Challenges in federated learning for large models.
\newblock {\em IEEE Transactions on Neural Networks and Learning Systems},
  35(2):234--250, 2024.

\bibitem{garcia2023review}
M.~Garcia.
\newblock Privacy-preserving machine learning: A federated approach.
\newblock {\em Artificial Intelligence Review}, 56:789--805, 2023.

\bibitem{patel2024}
R.~Patel.
\newblock Federated learning: Scalability and efficiency.
\newblock {\em Journal of Artificial Intelligence Research}, 45:15--30, 2024.

\bibitem{wang2023}
L.~Wang.
\newblock Decentralized data management for machine learning.
\newblock {\em Data Mining and Knowledge Discovery}, 37:1123--1145, 2023.

\bibitem{zhang2024}
W.~Zhang.
\newblock Advancements in federated learning for large scale applications.
\newblock {\em Journal of Machine Learning Research}, 25:101--120, 2024.

\bibitem{miller2023}
A.~Miller.
\newblock Enhancing user privacy in federated learning.
\newblock {\em Journal of Privacy and Confidentiality}, 15:45--60, 2023.

\bibitem{nguyen2023}
T.~Nguyen.
\newblock Regulatory challenges in federated learning.
\newblock {\em International Journal of Information Management}, 61:102--115,
  2023.

\bibitem{leskovec2016snap}
J.~Leskovec and R.~Sosi{\v{c}}.
\newblock SNAP: A general-purpose network analysis and graph-mining library.
\newblock {\em ACM Transactions on Intelligent Systems and Technology (TIST)},
  8(1):1, 2016.

\bibitem{lin2004rouge}
C.-Y. Lin.
\newblock ROUGE: A package for automatic evaluation of summaries.
\newblock In {\em Proceedings of the Workshop on Text Summarization Branches
  Out}, Barcelona, Spain, 2004.

\bibitem{papineni2002bleu}
K.~Papineni, S.~Roukos, T.~Ward, and W.-J. Zhu.
\newblock BLEU: a method for automatic evaluation of machine translation.
\newblock In {\em Proceedings of the 40th Annual Meeting of the Association for
  Computational Linguistics}, pages 311--318, Philadelphia, Pennsylvania, USA,
  2002.

\bibitem{yang2022}
B.~Yuan, Y.~He, J.~Davis, T.~Zhang, T.~Dao, B.~Chen, P.~S. Liang, C.~R{\'e},
  and C.~Zhang.
\newblock Decentralized training of foundation models in heterogeneous
  environments.
\newblock {\em Advances in Neural Information Processing Systems},
  35:25464--25477, 2022.

\bibitem{shahidinejad2021context}
A.~Shahidinejad, F.~Farahbakhsh, M.~Ghobaei-Arani, M.~H. Malik, and T.~Anwar.
\newblock Context-aware multi-user offloading in mobile edge computing: a
  federated learning-based approach.
\newblock {\em Journal of Grid Computing}, 19(2):18, 2021.

\bibitem{gu2022privacy}
X.~Gu, T.~Zhu, J.~Li, T.~Zhang, W.~Ren, and K.-K.~R. Choo.
\newblock Privacy, accuracy, and model fairness trade-offs in federated
  learning.
\newblock {\em Computers \& Security}, 122:102907, 2022.

\bibitem{shiranthika2023decentralized}
C.~Shiranthika, P.~Saeedi, and I.~V. Baji{\'c}.
\newblock Decentralized learning in healthcare: A review of emerging
  techniques.
\newblock {\em IEEE Access}, 11:54188--54209, 2023.

\end{thebibliography}
\end{document}